\documentclass{article}

\PassOptionsToPackage{numbers}{natbib}

\usepackage[final]{neurips_2024}

\usepackage{tabularx}
\usepackage{multirow}
\usepackage[utf8]{inputenc} 
\usepackage[T1]{fontenc}    
\usepackage{url}            
\usepackage{booktabs}       
\usepackage{amsfonts}       
\usepackage{nicefrac}       
\usepackage{microtype}      
\usepackage{xcolor}         
\usepackage{amssymb}
\usepackage{graphicx}
\usepackage{amsmath}
\usepackage{wrapfig}
\usepackage{subcaption}
\usepackage[most]{tcolorbox}
\usepackage{amsmath}
\usepackage{amsthm}
\usepackage{extpfeil}
\newtheorem{proposition}{Proposition}

\usepackage{algorithm}
\usepackage{algpseudocode}

\usepackage{xcolor}

\usepackage[colorlinks]{hyperref}

\def\tmp#1#2#3{%
  \definecolor{Hy#1color}{#2}{#3}%
  \hypersetup{#1color=Hy#1color}}
\tmp{link}{HTML}{800006}
\tmp{cite}{HTML}{2E7E2A}
\tmp{file}{HTML}{131877}
\tmp{url} {HTML}{8A0087}
\tmp{menu}{HTML}{727500}
\tmp{run} {HTML}{137776}
\def\tmp#1#2{%
  \colorlet{Hy#1bordercolor}{Hy#1color#2}%
  \hypersetup{#1bordercolor=Hy#1bordercolor}}
\tmp{link}{!60!white}
\tmp{cite}{!60!white}
\tmp{file}{!60!white}
\tmp{url} {!60!white}
\tmp{menu}{!60!white}
\tmp{run} {!60!white}

\usepackage[utf8]{inputenc}
\usepackage{geometry}
\usepackage{array}

\title{\textcolor[HTML]{4285F4}{S}\textcolor[HTML]{EA4335}{L}\textcolor[HTML]{FBBC04}{E}\textcolor[HTML]{34A853}{D}: Self Logits Evolution Decoding for Improving
Factuality in Large Language Models}

\author{
  Jianyi Zhang\textsuperscript{1},
  Da-Cheng Juan\textsuperscript{2}, Cyrus Rashtchian\textsuperscript{2},
  Chun-Sung Ferng\textsuperscript{2}, Heinrich Jiang\textsuperscript{2}, \\
  \textbf{Yiran Chen\textsuperscript{1}}
  \\
  \textsuperscript{1} Duke University,
  \textsuperscript{2} Google Research\thanks{Due to policy restrictions on the models used, this paper has been updated. For more detailed content, please refer to the previous version.}
  \\
  \href{https://jayzhang42.github.io/sled_page/}{\textcolor[HTML]{4285F4}{Project Website}}
}

\begin{document}

\maketitle

\begin{abstract}

Large language models (LLMs) have demonstrated remarkable capabilities, but their outputs can sometimes be unreliable or factually incorrect. To address this, we introduce \textbf{S}elf \textbf{L}ogits \textbf{E}volution \textbf{D}ecoding (SLED), a novel decoding framework that enhances the truthfulness of LLMs without relying on external knowledge bases or requiring further fine-tuning. From an optimization perspective, our SLED framework leverages the latent knowledge embedded within the LLM by contrasting the output logits from the final layer with those from early layers. It then utilizes an approximate gradient approach to enable latent knowledge to guide the self-refinement of outputs, thereby effectively improving factual accuracy. Extensive experiments have been conducted on established benchmarks across a diverse range of model families (Gemma, Qwen, Mixtral, gpt-oss) and scales (from 1B to 45B), including more advanced architectural configurations such as the mixture of experts (MoE). Our evaluation spans a wide variety of tasks and the results demonstrate that SLED consistently improves factual accuracy compared to existing decoding methods while maintaining natural language fluency and negligible latency overhead. Furthermore, it can be flexibly combined with other decoding methods to further enhance their performance.

\end{abstract}

\section{Introduction}

Large Language Models (LLMs) have achieved remarkable breakthroughs in recent years, demonstrating exceptional performance across various domains~\cite{anil2023palm,openai-chatgpt,GPT4report,team2023gemini}. However, a significant challenge associated with LLMs is their tendency to hallucinate or distort the truth, resulting in outputs that are not factual~\cite{huang2023survey,ji2023survey,zhang2023siren}. This issue of hallucination undermines the reliability and trustworthiness of LLMs in practical applications. A popular strategy for improving the LLM factuality involves refining the decoding process~\cite{shi2024thorough,welleck2024decoding}.\begin{wrapfigure}{r}{0.4\textwidth} 
 \vspace{-0.55cm}
  \centering
  \includegraphics[width=0.4\textwidth]{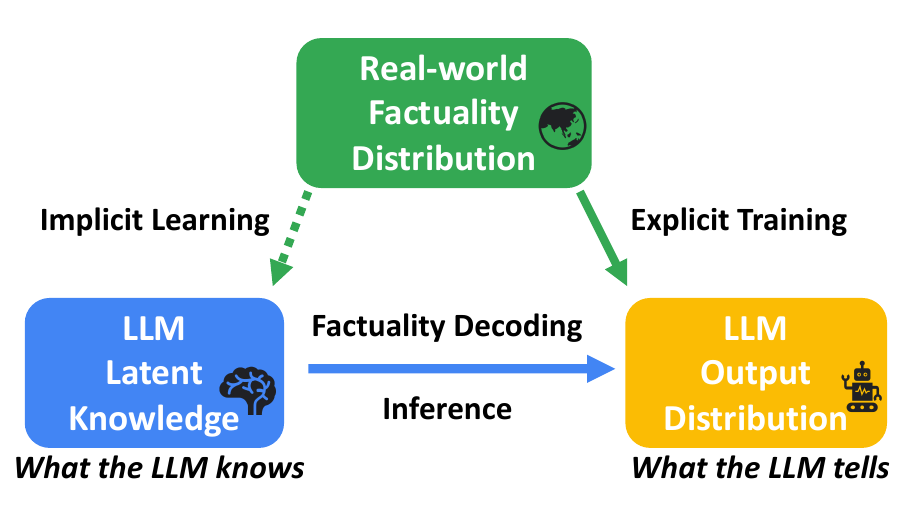} 
  \caption{Factuality decoding overview.} \label{fig:three-body}
\end{wrapfigure} Decoding focuses on how the model selects the next token during the generation process, which can significantly influence the factual accuracy of the output. The decoding methods can be cost-effective since (a) they do not rely on external knowledge and (b) no additional training is required. Furthermore, decoding methods can be synergistically combined with other techniques aimed at improving the LLM factuality, such as retrieving information from external knowledge bases~\cite{lei2023chain,lewis2020retrieval}, various fine-tuning strategies for better alignment~\cite{tian2024finetuning}, or ensemble learning methods~\cite{du2024improving}.

Recent studies~\cite{kadavath2022language,NEURIPS2023_ITI,saunders2022self,wang2020language} suggest that LLMs sometimes have learned the factual content based on extensive pretraining or fine-tuning, although they fail to produce the correct answer when a user queries the model. This has inspired the development of several factuality decoding methods \cite{chuang2024dola,NEURIPS2023_ITI,li2022contrastive,zhang2023alleviating} to reveal what the model implicitly "knows." Figure \ref{fig:three-body} summarizes the underlying mechanism of these factuality decoding methods. The LLMs' output distribution is derived by applying the softmax function to the output logits from the final layer. During the training phase, this distribution is optimized based on the real-world factuality distribution represented by the training dataset. However, during the inference phase, "what the LLM tells" might still contain factual errors, which implies a discrepancy between the output distribution and the real-world factuality distribution. While the real-world distribution remains inaccessible during the inference phase, the model's latent knowledge ("what the model knows") may have implicitly learned some factual content correctly during the training phase \cite{kadavath2022language,wang2020language}. Therefore, a key challenge for factuality decoding strategies lies in effectively harnessing the latent knowledge embedded within LLMs to refine the output distribution (logits) during inference.

\begin{figure*}[t!]
    \centering
    \vspace{-1.5cm}
    \includegraphics[width=0.93\textwidth]{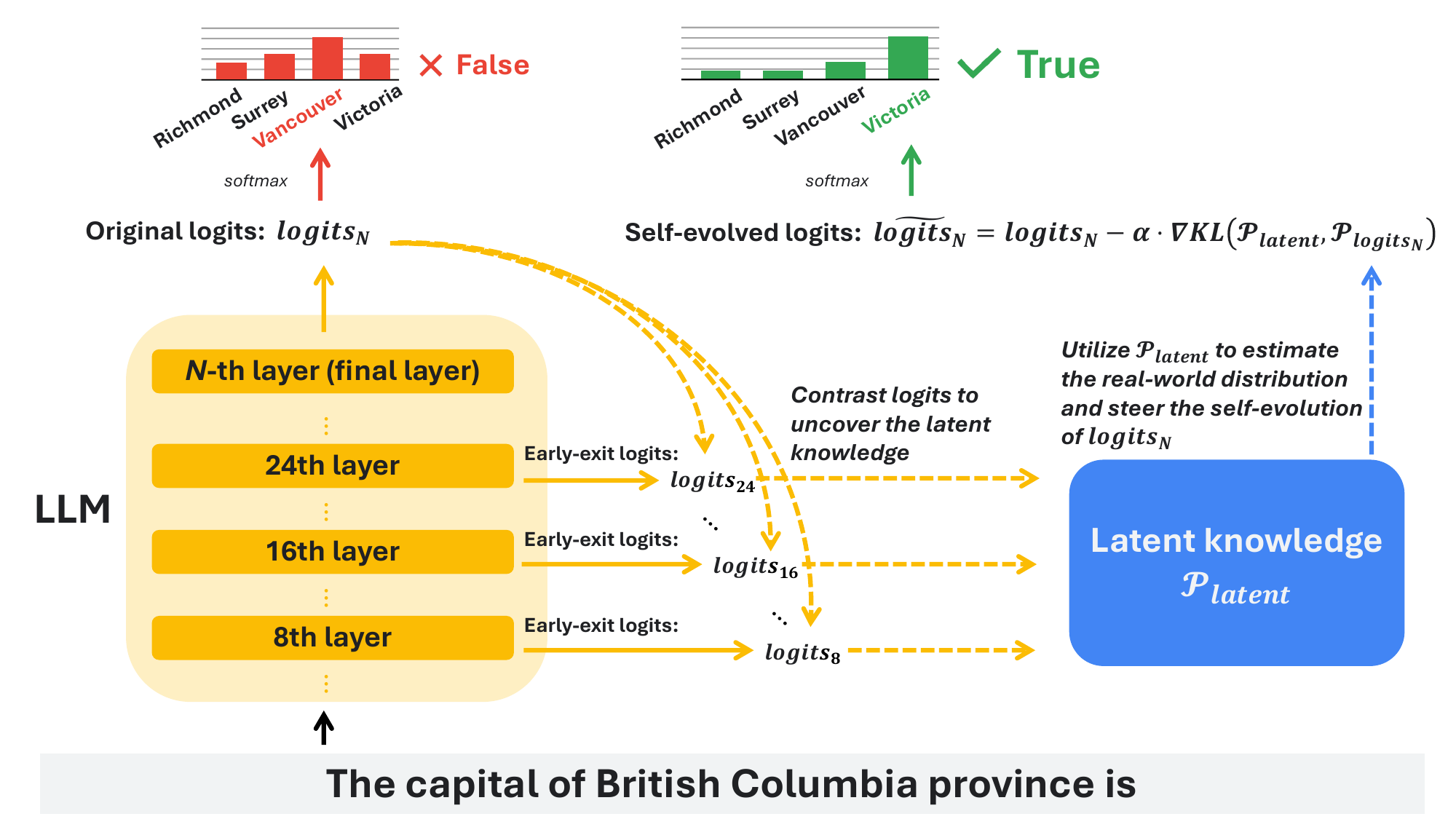}
    \caption{Illustration of our Self Logits-Evolution Decoding (SLED) workflow. }
    \label{fig:framework}
    \vspace{-0.5cm}
\end{figure*}

To address this challenge, we propose \textbf{S}elf \textbf{L}ogits \textbf{E}volution \textbf{D}ecoding (SLED), a novel factuality decoding approach that leverages the latent knowledge within LLMs by contrasting the final layer’s logits with early layers' logits. {During the decoding process, as LLMs progress from early to final layers, they progressively incorporate factual information stored in each layer into the output.} SLED tracks this evolution process to unearth the latent knowledge within LLMs, and enables the “self-evolution” of the output distribution further to align it more closely with real-world facts. Furthermore, our approach recognizes that the latent knowledge within LLMs, while valuable, may not always be perfect. Therefore, instead of simply replacing the original outputs with this latent knowledge, SLED integrates it into the original logits through an operation similar to “single-step gradient descent” over the output logits during the inference time. This operation minimizes the Kullback-Leibler (KL) divergence between the latent knowledge distribution and the output distribution, effectively balancing the two and mitigating potential drawbacks such as overfitting or biased outputs. Figure \ref{fig:framework} illustrates the SLED workflow, highlighting how  SLED optimizes the output logits, leading to a more factual output distribution. We evaluate SLED on various LLMs e.g., Gemma-3 \cite{gemmateam2025gemma3technicalreport}, Qwen-3 \cite{yang2025qwen3technicalreport}, GPT-OSS \cite{openai2025gptoss}) and benchmarks to demonstrate its state-of-the-art performance in layer-wise contrastive decoding methods. In summary, our main contributions are:
\begin{itemize}
    \item We propose SLED, a novel decoding method that aligns LLMs outputs with factual knowledge without requiring an external knowledge base or fine-tuning data.
    \item We conduct extensive experiments across a range of LLMs, with varying configurations and scales. The results demonstrate that SLED consistently improves factual accuracy on various tasks and benchmarks, including multiple-choice, open-ended generation, and chain-of-thought reasoning tasks.
    \item SLED can be flexibly integrated with other factuality decoding methods to enhance their effectiveness further.
    \item We provide a new interpretable perspective for understanding layer-wise contrastive decoding methods, paving the way for further developments in factuality decoding.
\end{itemize}

\vspace{-0.3cm}
\begin{algorithm}[t]
\caption{Self Logits Evolution Decoding} \label{algo1}
\begin{algorithmic}[1]

\State \textbf{Initialization:} LLM with $N$ layers, $inputs$, evolution rate $\alpha$, evolution scale $k>0$, $\eta \ll 0$, temperature parameter $\tau$, and the one-hoc vectors $\{\mathcal{P}_{e_i}\}$ defined in Section \ref{estimation_med}.

\State Feed the $inputs$ into the LLM to obtain the logits $\mathit{logits}_\mathit{n} = (\ell_{(1,\mathit{n})}, \ldots, \ell_{(d,\mathit{n})})$ and probabilities $\mathcal{P}_{\mathit{logits}_\mathit{n}} = (p_{(1,\mathit{n})}, \ldots, p_{(d,\mathit{n})}) = softmax(\mathit{logits}_\mathit{n}/\tau)$ at each layer $n$, where $n \leq N$.

\State Identify the tokens with the top-$k$ largest values in $\mathit{logits}_\mathit{N}$ and denote their indices by $I_k$.


\For{each early layer $n$, $(n<N)$}
    \State Compute differences for top-$k$ logits $\mathit{logits}_{\mathit{n}} - \mathit{logits}_{\mathit{N}}$.
    \State Calculate $m^{(n)}_i = \left[ \max \left( \text{CosSim} (\mathit{logits}_{\mathit{n}} - \mathit{logits}_{\mathit{N}}, \mathcal{P}_{\mathit{logits}_{\mathit{n}}} - \mathcal{P}_{e_i}), 0 \right) \right]^2, i \in I_k$.
\EndFor

\State Compute weighted average $m_i = \frac{\sum_{n=1}^N m^{(n)}_i}{\sum_{n=1}^N \sum_{j\in I_k} m^{(n)}_j}$ across different layers for each $i \in I_k$.

\For{each $i$ from $1$ to $d$}

\State Set $\tilde{\ell}_{(i,\mathit{N})} = {\ell}_{(i,\mathit{N})} - \frac{\alpha}{\tau} (p_{(i,\mathit{N})}-m_i )$ \textbf{if} {$i \in I_k$} \textbf{else} Set $\tilde{\ell}_{(i,\mathit{N})} = \eta \ll 0 $.

\EndFor

\State \textbf{Output:} The self-evolved logits are $\widetilde{\mathit{logits}}_\mathit{N} = (\tilde{\ell}_{(1,\mathit{N})},\ldots, \tilde{\ell}_{(i,\mathit{N})}, \ldots, \tilde{\ell}_{(d,\mathit{N})})$.

\end{algorithmic}
\end{algorithm}

\section{Self Logits Evolution Decoding}
A large language model,  equipped with $\mathit{N}$ layers and a vocabulary $\mathcal{V} = [v_1, v_2, \ldots, v_d]$, typically generates text in the next-token prediction fashion. For each given prefix, the
model computes the logits at the final ($\mathit{N}$-th) layer, $\mathit{logits}_\mathit{N} \triangleq (\ell_{(1,\mathit{N})}, \ell_{(2,\mathit{N})}, \ldots, \ell_{(d,\mathit{N})})$,  which are obtained by applying a linear transformation to the hidden states of the final layer, projecting the high-dimensional hidden state vectors onto the space of the vocabulary size. Subsequently, the output distribution $\mathcal{P}_{\mathit{logits}_\mathit{N}}$ at the final ($\mathit{N}$-th) layer for the next token is derived by applying softmax function on the logits, 
{\begin{align*}
    \mathcal{P}_{\mathit{logits}_\mathit{N}} \triangleq (p_{(1,\mathit{N})}, \ldots, p_{(d,\mathit{N})}) = softmax\left({\mathit{logits}_\mathit{N}}/{\tau}  \right),
\end{align*} }
where $\tau$ is the temperature parameter. Therefore, for each $p_{(i,\mathit{N})}~( 1 \leq i \leq d) $, we have \vspace{-3pt}
\begin{align*}
    p_{(i,\mathit{N})} = {\exp (\ell_{(i,\mathit{N})}/\tau)}/{S}, \ \text{where $\ S= \sum\nolimits_{j=1}^d \exp (\ell_{(j,\mathit{N})}/\tau).$} 
\end{align*}Similarly, we can also derive the logits from early layers by applying the same linear transformation mentioned above to their hidden states. For any early layer $n~(n < N$), we denote its logits as $\mathit{logits}_\mathit{n} \triangleq (\ell_{(1,\mathit{n})}, \ldots, \ell_{(d,\mathit{n})})$ and the corresponding distribution as $\mathcal{P}_{\mathit{logits}_\mathit{n}} \triangleq (p_{(1,\mathit{n})}, \ldots, p_{(d,\mathit{n})})$.


\subsection{Logits Evolution}\label{logits_evolution}

To improve factual accuracy, it is crucial that the correct token \(v_i\) receives a higher value of $\mathit{logits}_\mathit{N}$ to ensure a higher probability value $p_{(i,\mathit{N})}$ in the output distribution $\mathcal{P}_{\mathit{logits}_\mathit{N}}$. From a mathematical perspective, this means aligning the model's output distribution $\mathcal{P}_{\mathit{logits}_\mathit{N}}$  closely with the real-world factuality distribution $\mathcal{P}_{\mathit{real}}$. Specifically, we can formulate this goal as optimizing the following loss function $\mathcal{L}$ regarding the $\mathit{logits}$:
\begin{align}\label{training_evolution}
     \mathcal{L} ( \mathit{logits}) \triangleq  KL(\mathcal{P}_{\mathit{real}},\mathcal{P}_{\mathit{logits}}), \text{where}\ \mathit{logits} = (\ell_1, ...,\ell_d),\ \mathcal{P}_{\mathit{logits}}=softmax(\mathit{logits}/{\tau})
\end{align}
We describe the above optimization as \textbf{Logits Evolution}. Interestingly, the training of LLMs also aims at minimizing the divergence (typically the $\mathit{KL}$ divergence, as the training loss function is often the cross-entropy loss) between the ground truth $\mathcal{P}_{\mathit{real}}$ and the output distribution $\mathcal{P}_{\mathit{logits}_\mathit{N}}$. During the training phase, the logits evolution is driven externally by the real-world distribution $\mathcal{P}_{\mathit{real}}$ presented in the training dataset, and the corresponding solution is $\mathit{logits}=\mathit{logits}_\mathit{N}$. However, $\mathcal{P}_{\mathit{real}}$ is not accessible during the inference phase. To address this challenge, SLED utilizes the model's latent knowledge to estimate $\mathcal{P}_{{real}}$ and enables "self-evolution" of the logits. We denote the estimation as $\mathcal{P}_{\mathit{latent}}$ and the self logits evolution can be achieved by the following gradient-descent operation:
\begin{align}\label{optimized_lr_delta}
     \widetilde{\mathit{logits}}_\mathit{N} = \mathit{logits}_\mathit{N} - \alpha \cdot \nabla_{\mathit{logits}_\mathit{N}} KL(\mathcal{P}_{\mathit{latent}}, \mathcal{P}_{\mathit{logits}_\mathit{N}}). 
\end{align}
The parameter \(\alpha\), termed the \textbf{Evolution Rate}, governs the magnitude of adjustments applied to $\mathit{logits}_\mathit{N}$ in the direction of the gradient $\nabla_{\mathit{logits}_\mathit{N}} KL(\mathcal{P}_{\mathit{latent}}, \mathcal{P}_{\mathit{logits}_\mathit{N}})$. In the following Section \ref{method_starting_point} and \ref{estimation_med}, we discuss how we derive the $\mathcal{P}_{\mathit{latent}}$ as the estimation of the real-world distribution $\mathcal{P}_{\mathit{real}}$.

\subsection{Estimate $\mathcal{P}_{\mathit{real}}$ by Tracking the Logits Evolution Direction throughout Layers} \label{method_starting_point}

The core principle of our method involves leveraging the difference between each early layer's logits and the final layer's logit, $\mathit{logits}_\mathit{n} - \mathit{logits}_\mathit{N}$ to approximate the gradient of \(KL(\mathcal{P}_{\mathit{real}},\mathcal{P}_{\mathit{logits}})\) at \(\mathit{logits}= \mathit{logits}_\mathit{n}\). Then we estimate $\mathcal{P}_{\mathit{real}}$ based on this approximation. 

This is inspired by a new perspective of interpreting the training phase of LLMs as the evolution of logits described in Problem \ref{training_evolution}. As mentioned above, the solution derived by the training phase is the final layer's logits $ \mathit{logits}= \mathit{logits}_\mathit{N}$, since the final layer's $\mathit{logits}_\mathit{N}$ directly engage with the real-world distribution $\mathcal{P}_{\mathit{real}}$ through the loss function in training. This implies that we can generally consider the final logits $\mathit{logits}_\mathit{N}$ to be a better solution than the logits from an early layer $\mathit{logits}_\mathit{n}$, with \( KL(\mathcal{P}_{\mathit{real}},\mathcal{P}_{\mathit{logits}_\mathit{N}}) < KL(\mathcal{P}_{\mathit{real}},\mathcal{P}_{\mathit{logits}_\mathit{n}}) \). 
Based on this discussion, if we contrast the final layer's logits with the early layer's logits, we can consider the direction (orientation) of $\mathit{logits}_\mathit{n} - \mathit{logits}_\mathit{N}$ can approximately align with the direction of the gradient \(\nabla_{\mathit{logits}} KL(\mathcal{P}_\mathit{real}, \mathcal{P}_\mathit{logits}) |_{\mathit{logits}=\mathit{logits}_\mathit{n}}\). To further verify this motivation, we calculate the cosine similarity between $\mathit{logits}_\mathit{n} - \mathit{logits}_\mathit{N}$ and \(\nabla_{\mathit{logits}_\mathit{n}} KL(\mathcal{P}_\mathit{real}, \mathcal{P}_{\mathit{logits}_\mathit{n}})\) for thousands of tokens across different models in Figure \ref{fig:model_gradient_approx_cos}. We find that the majority of these values are positive, which means that the directions of these two vectors are close.

Hence, for each early layer \(n\), we propose to maximize the following function of cosine similarity and derive the \(\mathcal{P}^{(n)}_{\mathit{latent}}\) to estimate the $\mathcal{P}_{\mathit{real}}$.
\vspace{0.1cm}
\begin{align}\label{cos_approx_problem}
    \mathcal{P}^{(n)}_{\mathit{latent}} = \arg \max_{\mathcal{P}} \left( \text{CosSim} ( \mathit{logits}_\mathit{n} - \mathit{logits}_\mathit{N}, \nabla_{\mathit{logits}_\mathit{n}} KL(\mathcal{P}, \mathcal{P}_{\mathit{logits}_\mathit{n}}), 0 \right)
\end{align}
\vspace{-0.5cm}
\subsection{Achieving the Self Logits Evolution in Three Phases}\label{estimation_med}

\begin{figure}[t]
    \centering
    \vspace{-1cm}
        \includegraphics[width=1\textwidth]{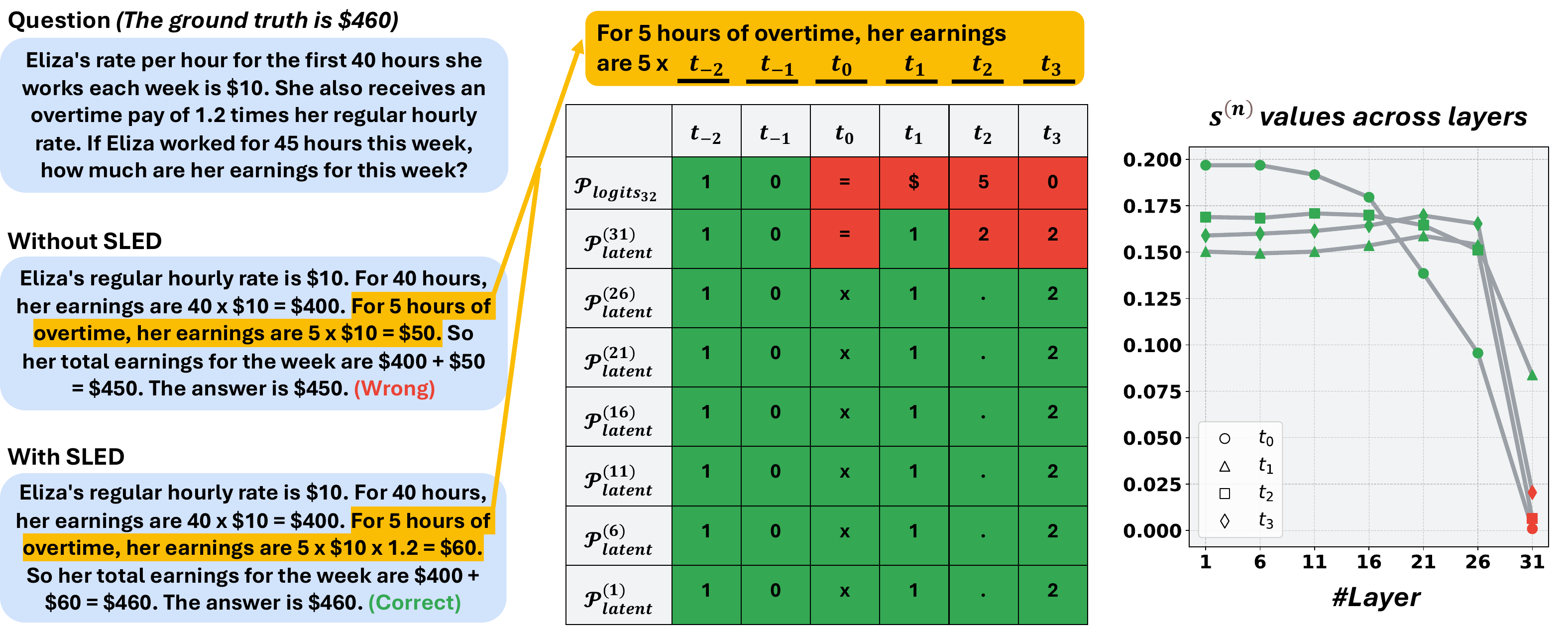}
        \vspace{-0.1cm}
            \caption{An example from GSM8K demonstrating SLED's mechanism. SLED derives the estimations $\mathcal{P}^{(n)}_{\mathit{latent}}$ by contrasting final layer's logits $\mathit{logits}_\mathit{N}$ with early layers' logits $\{\mathit{logits}_\mathit{n}\}$. We list the token with the highest probability value from the $\mathcal{P}^{(n)}_{\mathit{latent}}$ for different early layers. As shown, SLED downplays incorrect tokens by assigning lower weights $s^{(n)}$ to the corresponding $\mathcal{P}^{(n)}_{\mathit{latent}}$. Conversely, if the estimation is correct, the weights are relatively larger. The parameter evolution scale is set to 2.}        \label{fig:pipeline}
        \vspace{-0.2cm}
\end{figure}

Based on the above analysis, we can introduce the procedures of SLED: First, we estimate \(\mathcal{P}^{(n)}_{\mathit{latent}}\) for each early layer \(n\) using the gradient approximation in Section \ref{method_starting_point}. Subsequently, we apply a weighted average on $\{\mathcal{P}^{(n)}_{\mathit{latent}}\}$ across all early layers $n < N$ to derive \(\mathcal{P}_{\mathit{latent}}\), which serves as the final estimation of the real-world distribution. Finally, we apply \(\mathcal{P}_{\mathit{latent}}\) in Equation \ref{optimized_lr_delta} to facilitate the self-evolution of \({\mathit{logits}}_\mathit{N}\), thereby derive the updated logits, $\widetilde{\mathit{logits}}_\mathit{N}$.
\begin{align*}
{\mathit{logits}}_\mathit{n} - {\mathit{logits}}_\mathit{N} &\overset{\text{in direction}}{\approx} \nabla_{\mathit{logits}_\mathit{n}} KL(\mathcal{P}_\mathit{real}, \mathcal{P}_{\mathit{logits}_\mathit{n}}) \\
&\xRightarrow[\text{Estimate}]{\textcolor[HTML]{4285F4}{\text{Phase 1}}} \ \mathcal{P}^{(n)}_{\mathit{latent}} \ \xRightarrow[\text{Ensemble}]{\textcolor[HTML]{4285F4}{\text{Phase 2}}} \ \mathcal{P}_{\mathit{latent}} \ \xRightarrow[\text{Self-evolution in Eq \ref{optimized_lr_delta}}]{\textcolor[HTML]{4285F4}{\text{Phase 3}}} \widetilde{\mathit{logits}}_\mathit{N}
\end{align*}\paragraph{\textcolor[HTML]{4285F4}{Phase 1}:} An exhaustive search for an exact solution to the complex optimization problem (Equation \ref{cos_approx_problem}) is computationally impractical. We can reduce the solution space by the following. Suppose the real-world factuality distribution dictates that the next word to be generated is the \(i\)-th token \(v_i\) from the vocabulary $\mathcal{V}$. Thus \(\mathcal{P}_{\mathit{real}} = \mathcal{P}_{e_i}\), where \(\mathcal{P}_{e_i}\) represents a standard basis vector (one-hot vector) with the \(i\)-th component set to 1 and all other components set to 0. Then, we can simplify the aforementioned optimization problem by limiting the solution space to $\{\mathcal{P}_{e_i} \}_{i=0}^{d}$ and decide which token $i$ should be selected. The corresponding gradient when $\mathcal{P}=\mathcal{P}_{e_i}$ has the following formulation.
\begin{proposition}
The gradient of $KL(\mathcal{P}_{e_i},\mathcal{P}_{\mathit{logits}})$ at \(\mathit{logits} =\mathit{logits}_\mathit{n}\) is:
{\begin{align}
    \nabla_{\mathit{logits}_\mathit{n}} KL(\mathcal{P}_{e_i}, \mathcal{P}_{\mathit{logits}_\mathit{n}}) =  (\mathcal{P}_{\mathit{logits}_\mathit{n}} - \mathcal{P}_{e_i})/{\tau}  = \left(p_{(1,\mathit{n})},\ldots, p_{(i,\mathit{n})} - 1, \ldots, p_{(d,\mathit{n})}\right)/{\tau}
 \label{gradeint_p_latent}
\end{align}}
\end{proposition}\vspace{-0.2cm}
We calculate the cosine similarity between the gradient $\nabla_{\mathit{logits}_\mathit{n}} KL(\mathcal{P}_{e_i}, \mathcal{P}_{\mathit{logits}_\mathit{n}})$ and the difference $\mathit{logits}_\mathit{n} - \mathit{logits}_\mathit{N}$ for each token in the vocabulary $\mathcal{V}$. Then we select the \(\mathcal{P}_{e_{i^*}}\) of which the gradient is closest to $\mathit{logits}_\mathit{n} - \mathit{logits}_\mathit{N}$ as the estimation \(\mathcal{P}^{(n)}_{\mathit{latent}}\). Mathematically, this involves selecting \(i^*\) according to the following criterion
\begin{align*}
    i^* = \arg \max_{1\leq i \leq d} \, \bar{m}^{(n)}_i, \ \text{where} \ \bar{m}^{(n)}_i = \max \left( \text{CosSim} ( \mathit{logits}_\mathit{n} - \mathit{logits}_\mathit{N}, \mathcal{P}_{\mathit{logits}_\mathit{n}} - \mathcal{P}_{e_i} ), 0 \right),
\end{align*}
and adopting \(\mathcal{P}^{(n)}_{\mathit{latent}} = \mathcal{P}_{e_{i^*}}\) as the "hard estimation" of \(\mathcal{P}_{\mathit{real}}\). Drawing from the concept of hard and soft targets in label smoothing and knowledge distillation, we further extend it to the "soft estimation",
\begin{align*}
   \mathcal{P}^{(n)}_{\mathit{latent}} =   ( m^{(n)}_1, \ldots , m^{(n)}_i, \ldots, m^{(n)}_d)/{m^{(n)}}, \ \text{where} \ m^{(n)}_i = (\bar{m}^{(n)}_i)^2 \  \text{and} \ m^{(n)}={\sum\nolimits_{i=1}^d m^{(n)}_i}
\end{align*}
 We square $\{\bar{m}^{(n)}_i\}$ to moderately amplify their differences. Prior studies prove that soft targets usually offer stronger generalization capabilities, more information, and more robustness to noise than hard targets \cite{hinton2015distilling,muller2019does,ClassificationSoftRobust,zhang2021delving,zhang-etal-2023-reaugkd}. Hence, we adopt the soft estimation in lieu of the hard estimation.
\paragraph{\textcolor[HTML]{4285F4}{Phase 2}:} We ensemble \(\mathcal{P}^{(n)}_{\mathit{latent}}\) across all layers by computing a weighted average of the set \(\{\mathcal{P}^{(n)}_{\mathit{latent}}\}\) and adopt it as the final estimation of the $\mathcal{P}_{\mathit{latent}}$:
\begin{align*}
     \mathcal{P}_{\mathit{latent}} =  {\sum\nolimits_{n=0}^N s^{(n)} \mathcal{P}^{(n)}_{\mathit{latent}}}, \ \text{where} \ s^{(n)} = m^{(n)}/{(\sum\nolimits_{n=0}^N m^{(n)})}
\end{align*}
This estimation suggests that the weight $s^{(n)}$ of certain layer $n$ will be larger if the corresponding gradient approximation $\mathit{logits}_\mathit{n} - \mathit{logits}_\mathit{N}$ is more closely aligned with the gradients $\{\nabla_{\mathit{logits}_\mathit{n}} KL(\mathcal{P}_{e_i}, \mathcal{P}_{\mathit{logits}_\mathit{n}})\}$ for the tokens in the vocabulary. This in turn amplifies the influence of layer $n$ on the final estimation, which is a desirable effect in our method. Figure \ref{fig:pipeline} demonstrates that SLED can downplay incorrect tokens based on the gradient alignment.
One can further validate that for each component \(m_i\) in the final estimation \(\mathcal{P}_{\mathit{latent}} \triangleq (m_1, m_2, \ldots, m_d)\), the following relationship holds:
    $m_i = {\sum_{n=0}^N m^{(n)}_i}/{(\sum_{n=0}^N \sum_{j=1}^d m^{(n)}_j)}.$
    This property simplifies the description in Algorithm \ref{algo1}.      

\paragraph{\textcolor[HTML]{4285F4}{Phase 3}:} Applying $\mathcal{P}_{\mathit{latent}}$ in Equation \ref{optimized_lr_delta} enables us to derive the gradient necessary for steering the self-evolution on the final layer's logits $\mathit{logits}_\mathit{N}$.
\begin{proposition}\label{proposition2}
The gradient of $KL(\mathcal{P}_{\mathit{latent}},\mathcal{P}_{\mathit{logits}})$ at \(\mathit{logits} =\mathit{logits}_\mathit{N}\) is:
{\begin{align*}
    \nabla_{\mathit{logits}_\mathit{N}} KL(\mathcal{P}_{\mathit{latent}}, \mathcal{P}_{\mathit{logits}_\mathit{N}}) = ( \mathcal{P}_{\mathit{logits}_\mathit{N}} - \mathcal{P}_{\mathit{latent}})/{\tau}  = \left(p_{(1,\mathit{N})}-m_1,\ldots, p_{(d,\mathit{N})}-m_d\right)/{\tau}
\end{align*}}
\end{proposition}
\vspace{-0.2cm}
Then we can derive the self-evolved logits $\widetilde{\mathit{logits}}_\mathit{N}$ 
\begin{align} 
     \widetilde{\mathit{logits}}_\mathit{N} \triangleq (\tilde{\ell}_{(1,\mathit{N})},\ldots, \tilde{\ell}_{(i,\mathit{N})}, \ldots, \tilde{\ell}_{(d,\mathit{N})}),\
     \text{where} \ \tilde{\ell}_{(i,\mathit{N})} = {\ell}_{(i,\mathit{N})} - {\alpha} (p_{(i,\mathit{N})}-m_i )/{\tau}.
\end{align}

\subsection{Computational Complexity and Design Decisions}\label{sec:design_discussion}

For each layer, computing $\text{CosSim} ( \mathit{logits}_\mathit{n} - \mathit{logits}_\mathit{N}, \mathcal{P}_{\mathit{logits}_\mathit{n}} - \mathcal{P}_{e_i} )$ for every token $v_i$ in the vocabulary $\mathcal{V}$ needs $\mathcal{O}(d^2)$ operations. To reduce the computational complexity, we select only a subset $\mathcal{V}_{I_k}$, where the token $v_i \in \mathcal{V}_{I_k} $ has the top-$k$ highest logits in the final layer. In this scenario, we only initiate the self-evolution in Equation \ref{optimized_lr_delta} of the logits corresponding to these top-$k$ tokens. For the remaining tokens, which have lower probabilities, their logits are adjusted to a very lower numerical value, \textit{e.g.}, $-1000$. This strategy significantly reduces the computational complexity, while maintaining focus on the most relevant tokens. We name the parameter $k$, as \textbf{Evolution Scale}, since it determines the number of top-probability tokens active for self-evolution.

\textit{Q 2.1: Why SLED contrast the final layer with all the early layers, instead of picking one premature layer to contrast based on JSD?}

DoLa selects a subset of early layers to form a candidate set. Then it calculates the Jensen-Shannon Divergence (JSD) between the final layer and each layer in this set. Their strategy is to choose the layer with the highest JSD as the premature layer, and the chosen layer will be contrasted with the final layer to update probabilities. Obviously, if this strategy is reasonable, a larger candidate set should lead to a better choice of the premature layer and, consequently, enhanced overall performance. However, a paradoxical finding from their experimental results, which our tests also confirm, is that a larger candidate set for DoLa leads to decreased performance. As shown in ~\cite{chuang2024dola}, when the candidate set for DoLa ranged from 0-32 layers, the performance was inferior compared to a smaller set of 0-16 layers. This fundamental flaw indicates that selecting a good candidate set remains a challenge when applying DoLa. In contrast, our method does not face this concern as it applies an ensemble approach to all early layers. It is also important to note that our method works well even when only contrasting the final layer with part of the early layers, as demonstrated in the results, proving the robustness of our approach.


{\textit{Q 2.2: Why not use $\mathcal{P}_{\mathit{latent}}$ directly as the model’s output distribution? }}

It is crucial to understand that \(\mathcal{P}_{\mathit{latent}}\) is merely an estimation of the real-world distribution based on the model's latent knowledge instead of the exact \(\mathcal{P}_{\mathit{real}}\). Consequently, relying solely on \(\mathcal{P}_{\mathit{latent}}\), similar to DoLa, might lead to inaccuracies, as the latent knowledge can be imperfect. The original logits \(\mathit{logits}_\mathit{N}\) are still important as they are refined directly by real-world data during training. The evolution rate \(\alpha\) in Equation \ref{optimized_lr_delta}, serves to balance this trade-off, enabling a reciprocal enhancement between \(\mathcal{P}_{\mathit{latent}}\) and the original \(\mathit{logits}_\mathit{N}\).  More ablation studies are provided.


{\textit{Q 2.2: Considering that SLED adopts $\mathit{logits}_\mathit{n} - \mathit{logits}_\mathit{N}$ as the estimation of the gradient, why not directly apply it in Equation \ref{optimized_lr_delta}? }}

It is important to note that while $\mathit{logits}_\mathit{n} - \mathit{logits}_\mathit{N}$ is unconstrained, the gradients estimated in Equation \ref{proposition2} (e.g., $p_{(1,\mathit{N})}-m_1,\ldots, p_{(d,\mathit{N})}-m_d)$ are constrained within $[-1,1]$. Thus, direct substitution could lead to a mismatch in magnitudes and might also introduce unexpected noise. Proper normalization and subsequent aggregation of estimations from different layers are precisely what our method addresses in Section \ref{method_starting_point} and \ref{estimation_med}. Further analysis is provided.


\section{Experiments}
\label{section:experiments}
As a novel layer-wise contrastive decoding approach, we first benchmark SLED against the state-of-the-art approach DoLa \cite{chuang2024dola} across a diverse range of model families (Gemma, Qwen, Mixtral, gpt-oss) and model scales (from 2B to 45B), including the more advanced mixture of experts (MoE) architecture, as detailed in Section \ref{IOD-results}. The results showcase notable factuality improvements across a variety of tasks, including multi-choice, open-generation, and adaptations to chain-of-thought reasoning tasks. 



\subsection{Experimental Setup}
 
\begin{table}[]
\centering
\caption{Comparison on Gemma-3 model family. The best results are in bold for each dataset/metric. SLED outperforms DoLa and vanilla greedy decoding.} \label{main_gemma3}
\renewcommand{\arraystretch}{1.2} 
\resizebox{\textwidth}{!}{
\begin{tabular}{lllll|llllll}
\toprule
\multirow{2}{*}{Model} & \multirow{2}{*}{FACTOR} & \multicolumn{3}{c|}{TruthfulQA} & \multirow{2}{*}{Model} & \multirow{2}{*}{FACTOR} & \multicolumn{3}{c}{TruthfulQA} \\ \cline{3-5} \cline{8-10} 
                       &                         & \multicolumn{1}{c}{MC1}      & \multicolumn{1}{c}{MC2}      & \multicolumn{1}{c|}{MC3}      &                         &                         & \multicolumn{1}{c}{MC1}      & \multicolumn{1}{c}{MC2}      & \multicolumn{1}{c}{MC3}      \\ \midrule
Gemma-3-1B-PT             & 47.83                   & 32.28    & 63.22    & 30.98    & Gemma-3-12B-PT           & 66.67                   & 35.82    & 63.70   & \textbf{34.83}    \\
+DoLa                  & 12.63                  & 35.44    & 68.90    & 35.29    & +DoLa                   & 7.28                   & 33.54    & 66.62    & 34.25    \\
+SLED (ours)           & \textbf{63.29}          & \textbf{35.57}    & \textbf{71.79}    & \textbf{40.69}  & +SLED (ours)             & \textbf{74.25}          & \textbf{37.85}    & \textbf{67.05}    & 34.59 \\ \midrule
Gemma-3-1B-IT          & 37.17                   & 33.29    & 61.65    & 30.74    & Gemma-3-12B-IT        & 59.12                   & 39.49    & 63.62    & 37.89    \\
+DoLa                  & 16.50                   & 36.46    & 68.86   & 35.69    & +DoLa                   & 11.06                   & 35.19    & 67.53    & 35.04   \\
+SLED (ours)           & \textbf{55.04}          & \textbf{36.71}    & \textbf{71.79}    & \textbf{42.00} & +SLED (ours)             & \textbf{70.81}          & \textbf{46.33}    & \textbf{71.15}    & \textbf{40.90} \\ \midrule
Gemma-3-4B-PT            & 58.78                   & 33.80    & 64.12    & 32.64    & Gemma-3-27B-PT                  & 72.04                   & 36.20    & 61.39    & 34.95    \\
+DoLa                  & 9.25                   & 34.68    & 68.21    & 34.90    & +DoLa                   & 55.88                   & 31.65    & 65.03    & 32.72    \\
+SLED (ours)           & \textbf{69.94}          & \textbf{35.19}    & \textbf{69.25}    & \textbf{38.96} & +SLED (ours)             & \textbf{78.12}          & \textbf{38.48}    & \textbf{67.40}    & \textbf{35.65} \\ \midrule
Gemma-3-4B-IT         & 50.17                   & 36.96    & 58.42    & 34.13    & Gemma-3-27B-IT                & 66.00                   & 41.14    & 64.06    & 38.71    \\
+DoLa                  & 10.79                   & 36.08    & \textbf{68.59}     & 35.58    & +DoLa                   & 47.43                   & 33.16    & 65.93    & 33.63    \\
+SLED (ours)           & \textbf{66.15} & \textbf{45.19}          & 66.15   & \textbf{38.06}     & +SLED (ours)             & \textbf{73.75}          & \textbf{47.47}    & \textbf{73.58}    & \textbf{43.53} \\
\bottomrule
\end{tabular}
}
\vspace{-0.3cm}
\end{table}


\paragraph{Benchmarks} 
We compare our method with baselines on several multiple-choice. For multiple-choice question tasks, we use the TruthfulQA \cite{lin-etal-2022-truthfulqa} and FACTOR (Wiki) ~\cite{muhlgay2023generating} datasets to assess the LLMs' factuality in short-answer/long-paragraph scenario, respectively. 
\vspace{-0.2cm}

\paragraph{Models \& Baselines} We evaluate the performance of SLED on eight Gemma-3 models \cite{gemmateam2025gemma3technicalreport} (\{1B,4B,12B,27B\}-PT, \{1B,4B,12B,27B\}-IT), two Gemma-1 models (2B,7B), two MoE models (Mixtral-8$\times$7B, Mixtral-8$\times$7B-IT)~\cite{jiang2024mixtralexperts}, one Qwen-3 Model and one OpenAI gpt-oss-20b model. We adopt the following baselines: 1) standard decoding (greedy decoding or sampling depending on the tasks), 2) DoLa \cite{chuang2024dola}. 
\vspace{-0.4cm}

\paragraph{Metrics} We adopt the factual accuracy evaluation implemented in ~\cite{chuang2024dola} for multiple-choice tasks.
\vspace{-0.2cm}

\subsection{Evaluation on a Broad Range of LLM Benchmarks} \label{IOD-results}

The objective of these tasks is to employ decoding methods that enable LLMs to assign higher probabilities to correct completions/answers over incorrect alternatives. We demonstrate the effectiveness of SLED for both Short-Answer Factuality on the TruthfulQA and Long-Paragraph Factuality on the FACTOR dataset. For both DoLa and our SLED, we contrast the results from the final layer against all preceding layers. We randomly sample approximately 5\% of the data for validation regarding parameter selection. The results, as shown in Table \ref{main_gemma3}, indicate that SLED achieves superior outcomes in almost all metrics across 8 Gemma-3 models. Notably, SLED achieves better performance under the MC1/MC3 metrics on TruthfulQA, which are more sensitive to fluctuations and pose a greater challenge. For long sentences in FACTOR, our method shows improvements over baselines by 5-13\%. These results not only underscore the benefits of our method for factuality but also demonstrate its robustness across different lengths of text.

\subsection{Evaluation Across Diverse LLM Configurations}\label{sec:llama3}

As discussed above and shown in Table \ref{main_gemma3}, our method, SLED, demonstrates strong generalization capabilities across the Gemma-3 model family, proving robust from 1B to 27B model sizes. In Table \ref{gpt-oss}, we further showcase SLED's impressive performance on the other family models, such as Mixtral, Qwen-3, gpt-oss models, in terms of long paragraph factuality and short answer factuality. Interestingly, SLED is also applicable to the increasingly popular Mixture of Experts (MoE) architectures. These results confirm the exceptional adaptability of our method across various LLM configurations.

\begin{table}[htbp]
\centering
\caption{Using SLED with other LLM families also improves the factuality.} \label{gpt-oss}
\renewcommand{\arraystretch}{1.2} 
\resizebox{\textwidth}{!}{
\begin{tabular}{lllll|llllll}
\toprule
\multirow{2}{*}{Model} & \multirow{2}{*}{FACTOR} & \multicolumn{3}{c|}{TruthfulQA} & \multirow{2}{*}{Model} & \multirow{2}{*}{FACTOR} & \multicolumn{3}{c}{TruthfulQA} \\ \cline{3-5} \cline{8-10} 
                       &                         & \multicolumn{1}{c}{MC1}      & \multicolumn{1}{c}{MC2}      & \multicolumn{1}{c|}{MC3}      &                         &                         & \multicolumn{1}{c}{MC1}      & \multicolumn{1}{c}{MC2}      & \multicolumn{1}{c}{MC3}      \\ \midrule
Gemma-2B                  & 50.87                   & 23.38    & 37.16    & 17.42    & Mixtral-8$\times$7B           & 71.41                   & 35.13    & 49.98   & \textbf{34.17}    \\
+DoLa                  & 32.93                   & \textbf{26.07}    & 48.97    & 26.55    & +DoLa                   & 58.28                   & 32.44    & 35.91    & 33.68    \\
+SLED (ours)           & \textbf{57.05}          & 25.21    & \textbf{50.20}    & \textbf{26.94}  & +SLED (ours)             & \textbf{74.92}          & \textbf{35.86}    & \textbf{57.26}    & 32.96 \\ \midrule
Gemma-7B                 & 60.42                   & 31.58    & 47.63    & 22.75    & Mixtral-8$\times$7B-IT        & 70.51                   & 37.94    & 62.51    & 35.25    \\
+DoLa                  & 36.07                   & 25.21    & 43.14    & \textbf{26.13}    & +DoLa                   & 56.15                   & 32.19    & 39.17    & 33.76    \\
+SLED (ours)           & \textbf{65.56}          & \textbf{32.31}    & \textbf{49.88}    & 25.22    & +SLED (ours)             & \textbf{75.55}          & \textbf{41.73}    & \textbf{68.52}    & \textbf{37.70} \\ \midrule
gpt-oss-20b            & 41.12                   & 34.43   & 67.24    & 34.41    & Qwen-3-14B-Base                  & 57.69                   & 38.10    & 68.65    & 36.16    \\
+DoLa                  & 43.59                   & 28.99    & 61.72    & 30.33    & +DoLa                   & 58.42                   & 34.43    & 65.09    & 33.17    \\
+SLED (ours)           & \textbf{55.31}          & \textbf{36.71}    & \textbf{67.69}    & \textbf{35.26} & +SLED (ours)             & \textbf{64.09}          & \textbf{40.00}    & \textbf{68.99}    & \textbf{36.23} \\
\bottomrule
\end{tabular}
}
\end{table}

\section{Conclusion}

We introduced Self Logits Evolution Decoding (SLED), which is a new method to improve accuracy and factuality without requiring external knowledge (e.g., RAG) or fine-tuning (e.g., SFT). The key idea is to optimize the output logits based on the LLMs' latent knowledge to improve factuality during inference. On several datasets, SLED achieved the SOTA results, improving over the vanilla decoding and DoLa. SLED does not increase the inference time significantly, and it can be combined with other factuality decoding methods. For future work, it would be interesting to combine SLED with supervised fine-tuning methods, e.g., to adapt to other domains.

\section*{Acknowledgment}

This work was done when Jianyi Zhang was an intern at Google Research. In
addition, Jianyi Zhang and Yiran Chen disclose the
support from grants NSF CNS-2112562 and ARO W911NF-23-2-0224. We thank area chair and reviewers for their valuable comments.

\bibliographystyle{plainnat}
\bibliography{ref}

\end{document}